\pdfoutput=1

\documentclass[runningheads]{llncs}

\usepackage{amssymb}
\usepackage[usestackEOL]{stackengine}
\usepackage{amsmath}
\usepackage{enumitem}
\usepackage{listings}
\usepackage{afterpage}
\usepackage{mathrsfs}
\usepackage{amssymb}
\usepackage{mathtools}
\usepackage{adjustbox}
\usepackage{rotating}
\usepackage{makecell}
\usepackage{tikz}
\usepackage{url}

\usetikzlibrary{automata, matrix}
\DeclareMathAlphabet{\mathpzc}{OT1}{pzc}{m}{it}

\newcommand{\Cross}{{\tikz [x=1.2ex,y=1.2ex,line width=.2ex, red] \draw (0,0) -- (1,1) (0,1) -- (1,0);}}%

\usepackage{color}
\definecolor{dkgreen}{rgb}{0,0.6,0}
\definecolor{gray}{rgb}{0.5,0.5,0.5}
\definecolor{mauve}{rgb}{0.58,0,0.82}

\usepackage{graphicx}
\begin{document}
\title{Learning the Right Expansion-ordering Heuristics for Satisfiability Testing in OWL Reasoners}
\author{Razieh Mehri\inst{1} \and
Volker Haarslev\inst{1} \and
Hamidreza Chinaei\inst{2}}
\titlerunning{Learning Expansion-ordering Heuristics for OWL Satisfiability Testing}
\authorrunning{R.\ Mehri et al.}
\institute{Concordia University, Montr\'eal, Canada \\
\email{r\_mehrid@encs.concordia.ca, haarslev@cse.concordia.ca}\\
\and
Nuance Communications\\
\email{hamid.chinaei@nuance.com}}
\maketitle              % typeset the header of the contribution
\begin{abstract}
Web Ontology Language (OWL) reasoners are used to infer new logical relations from ontologies. While inferring new facts, these reasoners can be further optimized, e.g., by properly ordering disjuncts in disjunction expressions of ontologies for satisfiability testing of concepts. Different expansion-ordering heuristics have been developed for this purpose. The built-in heuristics in these reasoners determine the order for branches in search trees while each heuristic choice causes different effects for various ontologies depending on the ontologies' syntactic structure and probably other features as well. A learning-based approach that takes into account the features aims to select an appropriate expansion-ordering heuristic for each ontology. The proper choice is expected to accelerate the reasoning process for the reasoners. In this paper, the effect of our methodology is investigated on a well-known reasoner that is JFact. Our experiments show the average speedup by a factor of one to two orders of magnitude for satisfiability testing after applying learning methodology for selecting the right expansion-ordering heuristics.
\end{abstract}
\section{Introduction}
Satisfiability testing is an integral part of many reasoning tasks in Description Logic (DL) reasoners. Satisfiability tests generally contain non-deterministic choices. Different orders of choosing disjuncts in non-deterministic expansions have resulted in different speed values for reasoning tasks; hence, learning the right order for disjuncts in each expansion level is an open issue. 
\\A well-known decision-based method that is applied by SAT solvers involves the application of different branching heuristics in order to determine which concept (or branch) must be chosen when dealing with disjunctions \cite{marques1999impact}. Unlike SAT solvers, DL reasoners are designed to handle more than propositional satisfiability. Since search strategies and branching heuristics are shown to be very effective for SAT and Satisfiability Modulo Theory (SMT) based problems \cite{cimatti2013mathsat5}, encoding DL expressions to SMT might yield a potential solution. However, the encoded results might become very large and thus intractable for real-world ontologies \cite{gasse2009expressive}. 

While in propositional logic every concept is considered an atomic concept, in DL expressions, concepts are rather rich and may have complicated structures. In this case, applying branching heuristics such as MOMS similar to SAT solver (for DL reasoners, in every tree level) are not equivalently effective. The reason, given in Chapter 9 of the Description Logic Handbook, is that these methods do not take into account the dependencies from other branches and they could even reduce the reasoner performance \cite{baader2003description}. However, to enhance the effects of backjumping that is an optimized form of backtracking \cite{baker1994hazards}, one could select the branches that contain concepts with old dependencies, as suggested in Chapter 7 of Horrocks's Ph.D.\ thesis \cite{horrocks1997optimising}. Nonetheless, this approach could not boost up the performance as much either. Further, a learning-based heuristic technique was proposed in order to reduce the expansion of the disjuncts with a clash. It uses the characteristics of the already expanded clash-free disjuncts for detecting and prioritizing the not yet expanded clash-free disjuncts \cite{sirin2006wine}. The technique is mainly effective for ontologies with a large number of nominals.  

Since optimizing the expansion of each disjunction can dramatically speed up the reasoner performance, designing similar heuristic strategies that can benefit satisfiability testing for DL reasoners is both required and attainable. Hence, a new heuristic strategy for ordering branches (sub-concepts) is introduced in \cite{tsarkov2005ordering}. The suggested heuristics take into account the metrics of the concepts (or sub-concepts) such as frequency of the concepts in the ontology, size of the concepts and depth of their quantifiers. These metrics presumably play a major role for heuristic decisions in satisfiability tests of the reasoner. Therefore, in this paper, we learn to choose among these built-in heuristics that are developed for the inputs with more than just propositional logic that is $\mathcal{SHOIQ}$. The aim is to at least avoid the timeout termination of the reasoning process for ontologies by learning to choose the right heuristics.

The remainder of this paper is organized in the following order: In the next section related work is discussed. Later, preliminaries are given for the background knowledge on DL. The details of the customizable expansion-ordering heuristics that are suggested for reasoners (specifically FaCT++) are explained in Section 4. In Section 5, our learning methodology for selecting among those heuristics is presented. In Section 6, we discuss the experiments and finally in Section 7, we conclude and present future work.
\section{Related Work}
Machine learning aids with effective and successful decision-making in different non-deterministic states related to reasoning about ontologies, e.g., assigning the most optimal DL reasoner to a specific ontology based on the ontology's features \cite{pan2018predicting}.
Ontology features play a pivotal role in predicting reasoner performance. Many works have revealed the impact of Ontology features on machine learning solutions for reasoners \cite{kang2012predicting,alaya2015makes}. Ontologies are very complex and variant, i.e., choosing relevant features are critical when it comes to increasing the prediction accuracy for different ontology learning applications \cite{sazonau2014predicting}. Additionally, Automated Theorem Prover (ATP) frameworks such as E-MaLeS proposed machine learning-based feature tuning \cite{kuhlwein2013males}; however, depending on the specification of ontology learning problems, these frameworks may or may not be generalized. 

Selecting suitable heuristics by utilizing machine learning algorithms in satisfiability testing or similar tasks has substantially enhanced the runtime for variants of SAT, SMT and QBF solvers \cite {lagoudakis2001learning,samulowitz2007learning,kolb2018learning}. For resolution-based theorem provers, different statistical machine learning methods are used to predict the plausibility of search branches for faster query results \cite{goolsbey2017identifying}. 

In another work, inspired by the above-mentioned developments, branching heuristic selection has been applied for satisfiability testing for DL domain where a great improvement has been reported. However, the optimization solution is very limited and only applicable to propositional logic \cite{mehriapplying}. Moreover, in related work, we applied machine learning for the heuristic-based ToDo list optimization in DL reasoners. The learning-based approach identifies the best order of applying different rules during the reasoning process \cite{mehri2018optimizing}.

\section{Preliminaries}
DL is a formal logic-based language that represents knowledge in the form of ontologies \cite{baader2003description}. DL is composed of two levels: TBox and ABox. A TBox contains concepts that are related to object properties (roles) with restrictions over those object properties. An ABox contains assertions about concept instances. By reasoning over TBox and ABox, OWL reasoners reveal new information that is not explicitly observed in the ontology. Further, DL introduces various languages that each permits specific operators in their syntax. In the following, one of the languages called $\mathcal{SHOIQ}$ is introduced briefly.

\subsection{Brief Intro to Semantics and Syntax of $\mathcal{SHOIQ}$}
$\mathcal{SHOIQ}$ \cite{horrocks2007tableau} is a well-known expressive DL language that extends $\mathcal{SHIQ}$ by allowing nominals in its syntax. $\mathcal{S}$ allows the occurrence of transitive roles, $\mathcal{H}$ allows for role inclusion axioms, $\mathcal{O}$ for nominals, $\mathcal{I}$ for inverse roles, and $\mathcal{Q}$ for qualified cardinality restrictions. 

The formal semantics of a concept language is defined by an interpretation $\mathcal{I}=(\Delta ^ \mathcal{I},\cdot^\mathcal{I})$ in which $\Delta ^ \mathcal{I}$ is a domain and $\cdot^\mathcal{I}$ is an interpretation function. The domain is a non-empty set and the interpretation function maps: every instance name $a$ to an element ($a^\mathcal{I}\in\Delta^\mathcal{I}$); every atomic concept $A$ to a subset ($A^\mathcal{I}\subseteq\Delta ^ \mathcal{I}$) and every atomic role $R$ to a binary relation ($R^\mathcal{I}\subseteq\Delta^\mathcal{I}\times\Delta^\mathcal{I}$). A concept $C$ is satisfiable if there exists an $\mathcal{I}$ with $C^{\mathcal{I}} \neq \emptyset$.

DL axioms belong to one of the three types: TBox, RBox, and ABox. 

\begin{itemize}
\item
TBox is the terminological part of an ontology that is a finite set of General Concept Inclusion (GCI) axioms. A GCI axiom is of the form of $C\sqsubseteq D$ where $C$ and $D$ are concepts. It is satisfiable if there exists an $\mathcal{I}$ such that $C^{\mathcal{I}} \subseteq D^{\mathcal{I}}$. Then $C$ is called a subsumee of $D$ and $D$ a subsumer of $C$. An equivalence axiom can be written as two GCI axioms. Therefore, $C \equiv D$ also called definitional GCI can be written as $C \sqsubseteq D$ and $D \sqsubseteq C$. $C\sqsubseteq D$ is called a primitive GCI when $C$ is an atomic concept and $D$ is a concept description. 

\item
RBox is a finite set of rule inclusion axioms $R\sqsubseteq S$ and transitivity axioms ($Tra(R)$) where $R$ and $S$ are roles. An axiom $R\sqsubseteq S$ is satisfiable if there exists an $\mathcal{I}$ such that $R^{\mathcal{I}} \subseteq S^{\mathcal{I}}$. If $R\sqsubseteq S$ holds, then $Inv(R)\sqsubseteq Inv(S)$ also holds. The reflexive transitive closure over $\sqsubseteq$ is denoted by $\sqsubseteq^*$. $R$ is a simple role if no transitive role $S$ exists where $S\sqsubseteq^* R$. 

\item
ABox is the assertional part of an ontology that consists of a finite set of assertions about individuals (or instances of concepts) $C(a)$ where $a$ is an instance of $C$, i.e., $a^{\mathcal{I}} \in C^{\mathcal{I}}$, and roles where $R(a,b)$ states that $a$ and $b$ are individuals in the relationship $R$, i.e., ($a^{\mathcal{I}},b^{\mathcal{I}}) \in R^{\mathcal{I}}$.
\end{itemize}

A TBox $\mathcal{T}$ and its associated RBox $\mathcal{R}$ are satisfiable if there exists an $\mathcal{I}$ that satisfies all axioms in $\mathcal{T}$ and $\mathcal{R}$. An ABox $\mathcal{A}$ is satisfiable if there exists an $\mathcal{I}$ that satisfies all axioms in $\mathcal{T}$ and $\mathcal{R}$, and all assertions in $\mathcal{A}$.

Standard DL inference services that are relevant to our paper are concept satisfiability, TBox satisfiability (or consistency), i.e., $\top^{\mathcal{I}} \neq \emptyset$, and concept classification, which computes a taxonomy or subsumption hierarchy over all atomic concepts in a TBox $\mathcal{T}$.

For the sake of clarity, we only focus on the $\mathcal{ALC}$ subset of $\mathcal{SHOIQ}$. The DL $\mathcal{ALC}$ \cite{schmidt1991attributive} is an essential core of OWL and the basic expressive DL that covers disjunction, existential and universal quantifiers, etc. Hence, the syntax and semantics of operators, etc. of $\mathcal{ALC}$ are presented in Table \ref{tab:Infer}.

\begin{table}[t]
\centering
\caption{Syntax and semantics of $\mathcal{ALC}$}
\label{tab:Infer}
\begin{tabular}{|l|l|l|}
\hline 
\textbf{Operators, etc.}\hspace*{-1mm} & \textbf{Syntax}\hspace*{-1mm} & \textbf{Semantics}\\
\hline 
\hline 
Top &$\top$ &  $\top^\mathcal{I} = \Delta^\mathcal{I}$ \\
Bottom             & $\bot$&$\bot^\mathcal{I} = \emptyset$   \\
Concept  & $A$  & $A^{\mathcal{I}}\subseteq\Delta^{\mathcal{I}}$, $A$ is atomic \\
Negation  & $\neg C$  & $\Delta^{\mathcal{I}}\setminus C^{\mathcal{I}}$\\
Conjunct.  & $C\sqcap D$  & $C^{\mathcal{I}}\cap D^{\mathcal{I}}$\\
Disjunct.  & {\small{}$C\sqcup D$}  & {\small{}$C^{\mathcal{I}}\cup D^{\mathcal{I}}$}\\
Universal  & $\forall R.C$  & $\{x \,|\, \forall y\!:\! (x,y) \!\in\! R^{\mathcal{I}}\Rightarrow y \!\in\! C^{\mathcal{I}}\}$\hspace*{-3mm}\\
Existential & $\exists R.C$  & $\{x \,|\, \exists y\!:\! (x,y) \!\in\! R^{\mathcal{I}} \wedge  y \!\in\! C^{\mathcal{I}}\}$\\
\hline 
Role  & $R$  & $R^{\mathcal{I}}\subseteq\Delta^{\mathcal{I}}\times\Delta^{\mathcal{I}}$\\
\hline 
\end{tabular}
\end{table}

\section{Expansion-ordering Heuristics}
Like other DL reasoners, FaCT reasoner successors such as FaCT++\footnote{\url{https://bitbucket.org/dtsarkov/factplusplus/src}} and JFact\footnote{\url{https://github.com/owlcs/jfact/}}, are accompanied by various configuration settings that help to integrate optimization techniques for different reasoning tasks. One of the parameters (setting fields) in their configuration provides customizable expansion-ordering heuristics that determine the sorting order for concepts in disjunctions. These expansion-ordering heuristics are set for both satisfiability and subsumption tests. Since the focus of the current research is only satisfiability testing, for the rest of this paper, expansion-ordering heuristics are only referring to heuristics for satisfiability testing; however, both tasks are provided with the same heuristic options and definitions in the configuration setting. 

The configuration setting's field for satisfiability testing is called \texttt{orSortSat} and as highlighted in the JFact source code, it is set to specify the sort ordering of non-deterministic vertices in \texttt{DAG} structure for satisfiability tests. The \texttt{DAG} structure, which represents an input ontology, will be explained later in detail.

\begin{table}[t]
   \caption{Configuration labels} 
   \label{tab:confglabel}
   \centering % center the table
   \begin{tabular}{r|c|c|c|c|c|c|c|c|c|c|c|c} % alignment of each column data
   Config\# & 1 & 2 & 3 & 4 & 5 & 6 & 7 & 8 & 9 & 10 & 11 & 12 \\ 
   \hline
   \hline
   Labels & Sap & Sdp & Fap & Fdp & Dap & Ddp & San & Sdn & Fan & Fdn & Dan & Ddn 
   \end{tabular}
\end{table}

The configuration setting options for expansion-ordering heuristics are in form of a string ``\texttt{Mop}'' that is defined as follows: ``\texttt{M}'' is a sorting name character field, which has the options\footnote{For the sorting name character field, two other statistics such as the number of branches and generating rules in concepts have been recognized. However, they have not been suggested as options to be chosen from in the configuration setting of the source codes for neither JFact nor FaCT++ \cite{tsarkov2005ordering}. For the sake of simplicity, we do not use them here.}: \texttt{S} (stands for size), \texttt{D} (stands for depth), and \texttt{F} (stands for frequency). ``\texttt{o}'' is an ordering type character field, which has the options: \texttt{d} for descending and \texttt{a} for ascending. Finally, ``\texttt{p}'' is a preference character field, which is either \texttt{p} for preferring generating rules or \texttt{n} for not preferring generating rules. Rules such as $\exists$ and $\geqslant$ that add new nodes to a tableau are called generating rules. This leads to 12 heuristics for satisfiability testing as identified in Table \ref{tab:confglabel}. If ``\texttt{Mop}'' is \texttt{0} then there is no sorting order imposed. 

Before any reasoning process begins, JFact computes all of the necessary statistics about each concept (such as size, frequency, depth, etc) and stores them for later use that involves imposing orders for concepts in \texttt{DAG} vertices. \\
If no sorting order is provided (``\texttt{Mop}'' is \texttt{0}), the reasoner uses the default order based on the ontology's structure. The part of the reasoner's source code that determines the default sorting order in JFact looks like the following: 
\begin{lstlisting}
orSortSat_Default= isLikeGALEN ? "Fdn" : isLikeWINE ? "Sdp" : "Sap";
\end{lstlisting}
In the above code, \texttt{isLikeGALEN} are the ontologies with a GALEN \cite{rogers2001GALEN} structure look alike and \texttt{isLikeWINE} are the ones with a structure similar to the Wine and Food ontology\footnote{\url{https://www.w3.org/TR/owl-guide/}}. The above code indicates that the default \texttt{orSortSat} (sorting configuration for satisfiability testing) for \texttt{isLikeGALEN} ontologies is \texttt{Fdn} and for \texttt{isLikeWINE} ontologies is \texttt{Sdp} and for other ontologies is \texttt{Sap}. 

The default \texttt{orSortSat} in the source code are determined based on some experiments on few ontologies from \cite{tsarkov2005ordering} (and likely further unpublished experiments by the developers). The experiments in \cite{tsarkov2005ordering} contain a few ontologies from DL'98 and three specific ontologies with different ontology structures that are defined in the following:

\textbf{GALEN} (Generalized Architecture for Languages, Encyclopaedias, and Nomenclatures in medicine) is an ontology with medical terminologies and it contains 2749 classes with no instances and more than 400 CGIs.

\textbf{Wine and Food} is an ontology for wine classification with fewer classes compared to GALEN but more than 100 nominals and a more complex structure to reason about.

\textbf{DOLCE} (Descriptive Ontology for Linguistic and Cognitive Engineering) is an ontology with medium complexity and size \cite{gangemi2002sweetening}.

In the source code of the reasoners FaCT++ and JFact, GALEN look alike ontologies (\texttt{isLikeGALEN}) are the ones with the number of CGIs beyond a specific threshold but few or zero ABox assertions and the Wine look alike ontologies (\texttt{isLikeWINE}) are the ones with more than 100 nominals but fewer CGIs.

\subsubsection{Direct Acyclic Graph (\texttt{DAG}) structure for ontologies} 
\begin{figure}[ht]
\centering
  \includegraphics[scale=0.45]{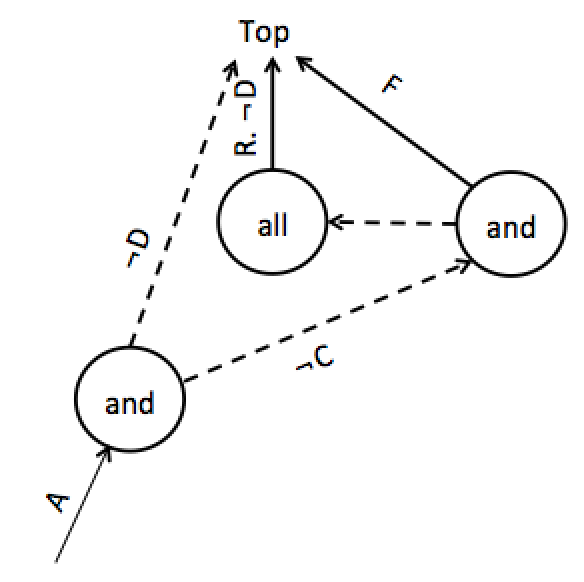}
  \caption{Direct Acyclic Graph (DAG) for TBox: $ C \sqsubseteq \exists R.D $ , 
$ C \sqsubseteq F$ 
and $ A \equiv C \sqcup D$ 
(edges determined by dash line will be negated before entering the next vertex)
}\label{dagT}
\end{figure}

JFact uses a \texttt{DAG} structure for representing ontologies. It transforms each ontology to a \texttt{DAG} structure in which the vertices of a graph are operators including: ``\texttt{and}''($\sqcap$), ``\texttt{all}''($\forall$), ``\texttt{at-most}''($\leq$) and the edges are operands. The \texttt{DAG} structure representation of ontologies helps with fast search and reasoning \cite{grigorev2013treasoner}.
\\
\\
For example, for the TBox containing:\\
$ C \sqsubseteq \exists R.D $ , 
$ C \sqsubseteq F$ 
and $ A \equiv C \sqcup D$ 
\\
Its equivalent \texttt{DAG} structure is shown in Fig \ref{dagT}.

JFact initially builds a \texttt{DAG} structure and before any satisfiability test happens it changes the orders of concepts in non-deterministic vertices of the graph based on the \texttt{orSortSat} field's value. 

To cover all possible non-determinism conditions of an ontology, the order of concepts in all of the vertices that starts with ``\texttt{and}'' are changing but considering that the most of satisfiability tests in ontologies are true, the disjunctions are mainly affected by this ordering modification as this is the main purpose of \texttt{orSortSat} heuristics \cite{tsarkov2005ordering}.

\section{Learning-based Approach to Select the Right Expansion-Ordering Heuristic}

Learning to make effective decisions in non-deterministic expansions are highly promising in satisfiability testing of reasoners. These decisions are particularly required for sorting of either atomic or non-atomic concepts in disjunctions.

As already noted, the default sorting order for expansion-ordering heuristics in the configuration setting of JFact is based on some initial assumptions that are obtained from past experiments that have analyzed a few syntactic features from a handful of ontologies, whereas other \textemdash possibly more important \textemdash features were disregarded. Moreover, as concluded in \cite{tsarkov2005ordering}, there is no general single sorting strategy that works best for all ontology types. Therefore, it is optimal to come up with a learning methodology that chooses the right sorting heuristic for each ontology based on its features such as both syntactic structure characteristics of the ontology as well as statistics from the \texttt{DAG} structure itself.

Building a multi-classification learning model that chooses from all 12 heuristics is not practical since the number of classes (heuristics) is too high and often leads to a low accuracy model. Instead, a more sensible solution is suggested in which for each heuristic a separate binary classification model is built that determines if that heuristic is the right choice. More details on the learning methodology are given in Section \ref{sec:experiments}.  

Support Vector Machine (SVM) is among one of the machine learning techniques that handles binary classification with a highly dimensional feature space effectively; therefore, it is used to build our highly dimensional model.

\subsubsection{Support Vector Machine (SVM)} 
\begin{figure}[htb]
\minipage{0.33\textwidth}
  \includegraphics[width=\linewidth]{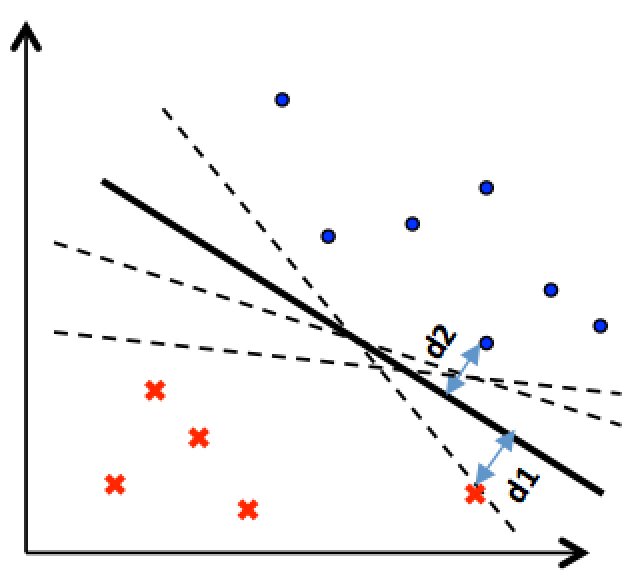}
  \caption{Optimal hyperplane for separating data points}\label{svmfig1}
\endminipage\hfill
\minipage{0.33\textwidth}
  \includegraphics[width=\linewidth]{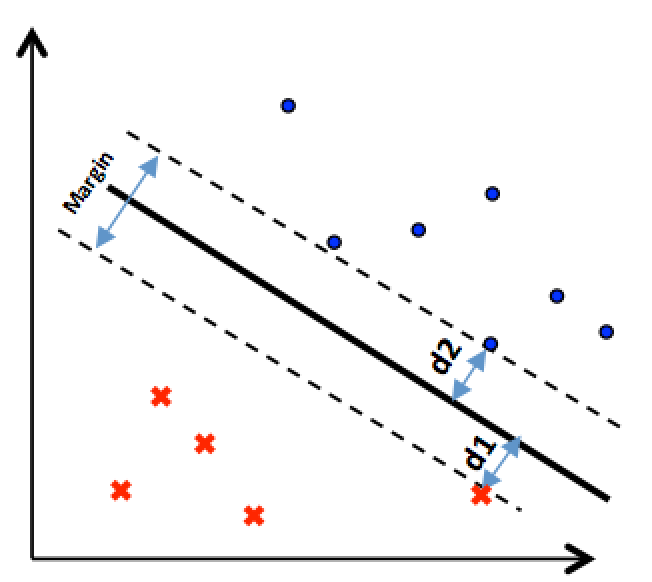}
  \caption{Hard margin}\label{svmfig2}
\endminipage\hfill
\minipage{0.33\textwidth}
  \includegraphics[width=\linewidth]{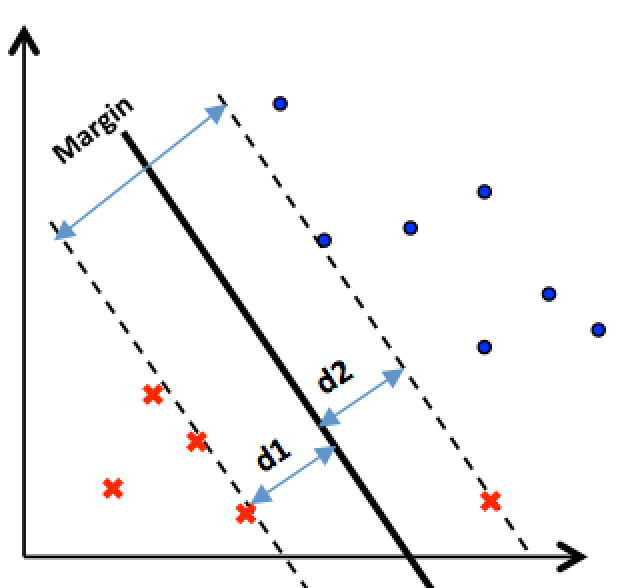}
  \caption{Soft margin}\label{svmfig3}
\endminipage\hfill
\end{figure}

is a supervised classifier technique that is best used for binary classification. SVM creates a decision boundary called hyperplane in order to separate data points (training data) into two different classes. Each data point belongs to one of the classes 1 or 2. As an example, Fig \ref{svmfig1} shows data points from two different classes 1 and 2 represented as \Cross \; and \tikz\draw[black,fill=blue] (0,0) circle (.5ex); ; respectively. There is an infinite number of hyperplanes that separate two classes; however, the objective is to find an optimal hyperplane that creates a maximum distance between the two closest data points from two classes. The optimal hyperplane in Fig \ref{svmfig1} is shown in bold. The margin is the location between the hyperplane and each data point shown in Fig \ref{svmfig2}. If the margin is small enough to separate the data points perfectly, it is called hard margin (Fig \ref{svmfig2}). However, there is a possibility that a model with 100\% accurate prediction cannot be generalized for unseen data. This issue is called overfitting of the model. To avoid overfitting, the margin should be softened (widened) to allow some misclassification as shown in Fig \ref{svmfig3}. This helps to create a more generalizable model.

\begin{figure}[htb]
\centering
  \includegraphics[scale=0.25]{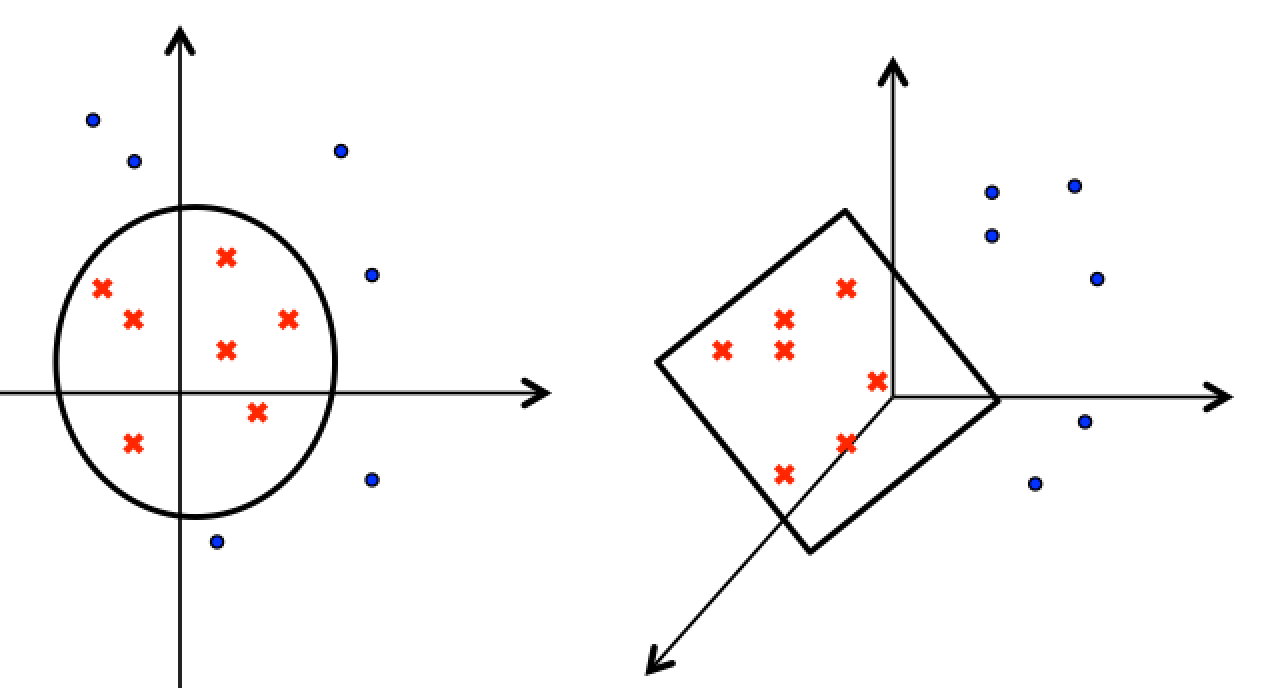}
  \caption{Mapping original data to a higher dimensional feature space}\label{svmfig4}
\end{figure}

SVM can handle the data points that are not linearly separable by using a non-linear hyperplane. In this case, the non-linear kernel functions are used to transform the original data point to a higher dimensional feature space to make them linearly separable in the new space as shown in Fig \ref{svmfig4}. Possible kernels are RBF (Radial Basis Function) and Polynomial \cite{cortes1995support}.

\subsection{Features}
Ontology features play a significant role in reasoning tasks that involve decision makings. Therefore, to implement a robust learning strategy, features are defined to cover the structural characteristics from ontology knowledge bases as well as the statistics from the \texttt{DAG} structure that are computed by JFact. Overall 39 features were used to build our model. A selection of the features that are believed to be the key ones in increasing the accuracy of our model is shown in Table \ref{tab:features}. 

\begin{table}[t]
\caption {Main features of ontologies for the built model} \label{tab:features} 
\centering
\begin{minipage}{.65\textwidth}
\begin{tabular}{l}
%\hline
\textbf{Ontology Type Metrics}\\ 
\hline
Number of nominals\\
Number of instances\\
Number of classes\\
Average population \\
Number of GCI\\
Number of generating rules ($\exists$ and $\geqslant$)\\
TBox ratio\\
RBox ratio\\
ABox ratio\\
Number of Object Properties\\
Number of Inverse Object Properties\\
Number of Subclasses\\
Number of Equivalent Classes\\
Number of Disjoint Classes
\end{tabular}

\smallskip

\begin{tabular}{l}
%\hline
\textbf{Non-deterministic Vertices' Metrics}\\ 
\hline
Number of vertices\\
Average of the size of vertices' sub-concepts' averages\\
Average of the depth of vertices' sub-concepts' averages\\
Average of the frequency of vertices' sub-concepts' averages\\
Maximum number of children in vertices \\
Average number of children in vertices \\
Number of positive concepts is vertices\\
Number of negative concepts is vertices\\
Ratio of positive concepts to all concepts in vertices\\
Ratio of negative concepts to all concepts in vertices
\end{tabular}
\end{minipage}
\end{table}

As mentioned previously, rules such as $\exists$ and $\geqslant$ that add a new node to tableau are called generating rules, which are considered as one of the features in Table \ref{tab:features}.

The average population feature is defined as the ratio of instances to classes in an ontology. This and some other features such as GCI consider the synthetic structure of ontologies that determine the ontology types or their similarity to GALEN and Wine or other types of ontologies. 

\subsection{Feature Transformation}
Feature transformation techniques increase the accuracy of the learning model by creating new features based on the original features. These techniques may include standardization as well as dimensional reduction techniques. 

\subsubsection{Standardization} helps to scale the data point, i.e., forcing them to have the same scale by assigning their mean to zero and standard deviation to one.

Dimensional reduction techniques are used to reduce the high dimension of feature spaces. One of the well-known feature reduction techniques is called PCA that transforms features to a set of new features that are non-linearly correlated \cite{wold1987principal}.

\subsubsection{Principal component analysis (PCA)}

When data is complex with too many correlated features, in order to reduce the difficulties of data, one may consider analyzing the data and its features. However, it is nearly impossible to interpret all the patterns that result from analyzing each feature against other features. PCA is a statistical technique used to reduce the dimensionality of features in order to reduce data complexity while keeping as much information as possible of the original data. In most cases, PCA increases the accuracy of models by creating a new combination of features. In this case, PCA uses a linear combination of features with different weights. Each principal component is a linear combination of features with maximum variance but uncorrelated with other principal components. Interpreting these principal components against each other is less expensive. Standardization as described above is usually an important pre-step for PCA that helps to highlight useful features.

\subsection{Feature Selection}

For our purpose of learning the right heuristic for satisfiability testing in reasoners, it is not easy to predict what features, among the 39 features, contribute more to the accuracy of the built model unless a feature selection technique is used. Feature selection techniques give scores to features based on their importance and finally select the most important features to build a model. The more a feature is correlated to the class (heuristics) prediction the more influence it has. One of the well-known feature selection techniques is Mutual Information.

\subsubsection{Mutual Information (MI)}

Mutual information measures the association between features and classes and scores the features accordingly.
\\
Mutual information formula is as follows:
\begin{equation}
\begin{aligned}
MI(F,C)= H(F)+H(C)-H(F,C)= H(F)+H(F|C)=\\
{\sum_{f\in F}^{}}{\sum_{c\in C}^{}} p(f,c) \log _{} {\frac{p(f,c)}{p(f)p(c)}}
\end{aligned}
\end{equation}
where $F$ is a set of features and $C$ is a set of classes. $H(X)$ is a marginal entropy for discrete random variable $X$, which calculates the amount of not known knowledge about $X$. $H(X|Y)$ and $H(X,Y)$ are the conditional entropy for variable $Y$ and joint entropy between $X$ and $Y$; respectively \cite{cover1991entropy}.

\section{Experimental Evaluation}
\label{sec:experiments}

This section reports on the impact of using the right expansion ordering heuristics on the performance of JFact. The reasoning task to be evaluated is satisfiability testing. Satisfiability testing for concepts of ontologies is preceded by a task that is consistency checking of ontologies. For some ontologies, consistency checking time is also affected by the expansion orderings. 

For simplicity, in this section, the expansion ordering heuristics are referred to as configurations. In Table \ref{tab:confglabel}, a unique configuration number is assigned to each of the expansion ordering heuristics.

To perform our experiments, ontologies from the ORE 2014 competition\footnote{\url{http://dl.kr.org/ore2014/ontologies.html}} dataset are used.

The ontologies collected from ORE 2014 must be eligible for building the learning model, i.e., the ones that are affected by the expansion ordering heuristics of satisfiability testing due to the existence of non-deterministic situations.

Based on that, initially, we were able to export around 3000 ontologies that contain non-deterministic situations and they must, at least, be in the form of a base logic ($\mathcal{ALC}$).
Additionally, two more eligibility criteria are considered for collecting the training data.

First, if for a single ontology, all the 12 configurations lead to a timeout, then that ontology is discarded because it is not clear if the ontology is affected by those heuristics. The considered timeout, for this experiment, is 500,000 milliseconds (8.33 minutes). Since it is an exhausting and non-practical process to obtain the exact termination runtime for ontologies that lead to a timeout, no number is assigned to them.

Second, if an ontology has very close runtimes from 12 configurations one may doubt its eligibility. The reason is that the difference could be just a runtime overhead from the system. To ensure this, multiple runtimes are examined and if for all runtimes the maximum and minimum runtimes belong to the same configurations, those ontologies are allowed in the training data because it is obvious that the heuristics are effective for that ontology.
Moreover, the ontologies that are inconsistent or result in similar issues are discarded too.

Considering all the above conditions for all ontologies from ORE 2014 dataset, only 143 from the dataset become qualified for training data collection where 25\% is dedicated to the test part. 

\begin{table}[t]
   \caption{Standard deviation and mean values for the threshold (numbers in milliseconds)} 
   \label{tab:std}
   \centering % center the table

   \begin{tabular}{c|c|c|c|c|c|c} % alignment of each column data
   Config\# & 1 & 2 & 3 & 4 & 5 & 6 \\ 
   \hline
   \hline
   std & 45240&	41093&	44969&	30237&	39744&	31692 \\
   mean & 90670&	74557&	84560&	70336&	82350&	68626 \\
   std+mean &135910	&115650	&129529	&100573	&122094	&100318 
   \end{tabular}
   
   \medskip
   
   \begin{tabular}{c|c|c|c|c|c|c} % alignment of each column data
   Config\# & 7 & 8 & 9 & 10 & 11 & 12 \\ 
   \hline
   \hline
   std& 70340&	48232&	45791&	31736&	52981&	39864 \\
   mean &	119611&	83761&	85775&	71612&	99992&	71706 \\
   std+mean	&189951	&131993	&131556	&103348	&152973	&111570
   \end{tabular}
\end{table}

As stated previously, for each configuration (heuristic), a binary classification model is built. For each binary model of a configuration, there are two labels: ``Good'', which indicates that a heuristic is the right choice and ``Bad'', which indicates the wrong choice. To determine a threshold for identifying the ``Good'' and ``Bad'' labels, the average of the mean plus standard deviation of the training data for all those configurations are considered. Table \ref{tab:std} shows the mean, standard deviation and their sum for each configuration. The average of the mean plus standard deviation for all configurations is about 127,122 (about 2 minutes) that indicates the threshold.

Considering a third label (as ``Ok''); which is assigned to the runtimes between threshold and timeout; may help with not only the right but also probably the best heuristic selection (the heuristic with the least runtime). Since defining more than two labels decreases the accuracy of the model, i.e., we prefer not to define it for the present.

To build each binary classifier, a 10-fold cross-validation SVM classifier is applied on the training data. To achieve the model with the highest accuracy, a grid search is conducted in order to find the best parameters such as the number of features in mutual information feature selection, number of PCA components, SVM kernels, etc. For configurations 7 and 8, SVM with RBF kernel outperformed linear SVM.

\begin{table}[t]
   \caption{Priorities of configurations for ``Good'' labels based on 10-fold cross-validation accuracy} 
   \label{tab:priority}
   \centering % center the table
   \begin{tabular}{r|c|c|c|c|c|c|c|c|c|c|c|c} % alignment of each column data
   Config\# & 1 & 2 & 3 & 4 & 5 & 6 & 7 & 8 & 9 & 10 & 11 & 12 \\ 
   \hline
   \hline
   Accuracy & 95\% & 83\% & 89\% & 89\% & 97\% & 91\% & 86\% & 82\% & 87\% & 93\% &  91\% & 84\% \\
   \hline
   Priority & 2 & 11 & 6 & 7 & 1 & 4 & 9 & 12 & 8 & 3 & 5 & 10
   \end{tabular}
\end{table}

The accuracy of each configuration and its priority based on its accuracy are given in Table \ref{tab:priority}. If two configurations hold an equal accuracy, then the priorities are given from left to right, e.g. configuration 3 has a higher priority than configuration 4. The priorities are used to determine which heuristic is a better choice for an input test ontology. 

\begin{sidewaystable}[tp]
%\thisfloatpagestyle{empty}
\fontsize{6}{9}\selectfont
%\centering
\caption{Runtimes (in milliseconds) for 35 satisfiability test cases selected from the ORE 2014 dataset (a bold font indicates the runtimes for the configuration chosen by our learning-based reasoner and falsely predicted labels are marked by `$^*$')} 
\label{tab:test}
\begin{adjustbox}{width=1\textheight,center=\textwidth}
\begin{tabular}{|l|c|c|c|c|c|c|c|c|c|c|c|c|c|c|}
  \hline
 Sample & Config 1 & Config 2 & Config 3 & Config 4 & Config 5 & Config 6 & Config 7 & Config 8& Config 9& Config 10& Config 11& Config 12 && JFact Std. Config\\
  \Xhline{2\arrayrulewidth}
   1 & 219016 & 98874 & 215667 & 90479 & 219501 & 104240 & 225228 & 98056 & 212528 & \textbf{87984} & 214417 & 109944 && 206076 \\
   \Xhline{0.5\arrayrulewidth}
   2 & 50136 & 1701 & 51144 & 1658 & \textbf{50219} & 1791 & 50967 & 1786 & 50037 & 1683 & 49214 & 1809 && 1762 \\
   \Xhline{0.5\arrayrulewidth}
   3 & 250 & 238 & 211 & 240 &  \textbf{219} & 225 & 258 & 3714 & 240 & 228 & 228 & 8511 && 210\\
   \Xhline{0.5\arrayrulewidth}
   4 & 791 & $1625^*$ & 641 & 218 & \textbf{807} & $1487^*$ & 818 & 2210 & 606 & 319 & 791 & 2245 && 771 \\
   \Xhline{0.5\arrayrulewidth}
   5 & $TO$ & 33833 & $TO$ & 28795 & $TO$ & 36608 & $TO$ & $TO$ & $TO$ & \textbf{22632} & $TO$ & $TO$ && $TO$\\
   \Xhline{0.5\arrayrulewidth}
   6 & 255 & 3607 & 588 & 620 & \textbf{266} & 662 & 289 & 3650 & 590 & 641 & 271 & 615 && 255 \\
   \Xhline{0.5\arrayrulewidth}
   7 & $4041^*$ & $3082^*$ & \textbf{4111} & $3138^*$ & $4201^*$ & $3243^*$ & $3088^*$ & 3081 & 4679 & $3085^*$ & $4208^*$ & 3170 && 3929 \\
   \Xhline{0.5\arrayrulewidth}
   8 & 660 & $TO^*$ & $TO^*$ & 445 &  \textbf{875} & $TO^*$ & 907 & $TO^*$ & $TO^*$ & 435 & 896 & $TO^*$ && 703\\
   \Xhline{0.5\arrayrulewidth}
   9 & $TO^*$ & 234 & $TO^*$ & 242 & $\textbf{TO}^*$ & 240 & $TO^*$ & 244 & $TO^*$ & 250 & $TO^*$ & 257 &&  $TO$\\
   \Xhline{0.5\arrayrulewidth}
   10 & 10487 & 2356 & 5190 & 1954 & \textbf{10419} & 2429 & 12158 & $TO^*$ & 5231 & 1970 & 11371 & $TO^*$ && 10403 \\
   \Xhline{0.5\arrayrulewidth}
   11 & $TO$ & 13074 & $TO$ & 10867 & $TO$ & 15434 & $TO$ & 13233 & $TO$ & \textbf{10152} & $TO$ & $14901^*$ && $TO$\\
   \Xhline{0.5\arrayrulewidth}
   12 & 5156 & $25706^*$  & 12763 & 1850 & \textbf{5703} & 5693 & $TO^*$ & TO & 12882 & 1339 & 5257 & 5685 && 26578\\
   \Xhline{0.5\arrayrulewidth}
   13 & 52562 & 13425 & 49296 & 11156 & \textbf{50658} & 16650 & 55552 & 14108 & 48152 & 11475 & 50508 & 16532 && 54102 \\
   \Xhline{0.5\arrayrulewidth}
   14 & 8769 & 3797 & 8762 & 3815 & \textbf{8744} & 9739 & 8397 & 3693 & 8605 & 3182 & 8543 & 8816 && 8615\\
   \Xhline{0.5\arrayrulewidth}
   15 & 27569 & $30690^*$ & $30752^*$ & $24417^*$ & \textbf{26763} & $24450^*$ & $27193^*$ & 23954 & $28244^*$ & 24752 & 26981 & 24492 && 36137\\
   \Xhline{0.5\arrayrulewidth}
   16 & 64886 & 19113 & 60452 & 17049 & \textbf{61710} & 22812 & 69656 & 18843 & 60778 & 16559 & 63575 & 22548 && 67340 \\
   \Xhline{0.5\arrayrulewidth}
   17 & $\textbf{194303}^*$ & 31035 & 164548 & 14655 & 167601 & 22773 & 203327 & $72764^*$ & $187606^*$ & 14606 & 159787 & 74447 && 165275\\
   \Xhline{0.5\arrayrulewidth}
   18 & 263574 & 112662 & 258198 & 104167 & 270668 & 121018 & 282923 & 112816 & 253426 & \textbf{101449} & 264520 & 121824 && 270144\\
   \Xhline{0.5\arrayrulewidth}
   19 & 223 & 261 & 256 & 239 & \textbf{238} & 233 & 254 & 2002 & 254 & 230 & 234 & 8685 && 224\\
   \Xhline{0.5\arrayrulewidth}
   20 & 1081 & $TO$ & 6375 & 1118 & \textbf{2989} & 1826 & 1855 & $TO$ & 6329 & 877 & 2930 & 1870 && $TO$ \\
   \Xhline{0.5\arrayrulewidth}
   21 & 81995 & $317447^*$ & 61637 & $181266^*$ & \textbf{59849} & $328572^*$ & 84153 & $210318^*$ & 61372 & $177632^*$ & 60545 & $197244^*$ && 81484\\
   \Xhline{0.5\arrayrulewidth}
   22 & 1599 & $TO^*$ & 12796 & 1840 & \textbf{6939} & 4029 & 41723 & $TO$ & 13344 & 1183 & 6056 & 3896 && $TO$\\
   \Xhline{0.5\arrayrulewidth}
   23 & 399 & 755 & 467 & 585 & \textbf{547} & 539 & 429 & 2415 & 469 & 616 & 726 & 842 && 378 \\
   \Xhline{0.5\arrayrulewidth}
   24 & $TO$ & 60399 & $TO$ & 47739 & $TO$ & 49117 & $TO$ & $TO$ & $TO$ & \textbf{48104} & $TO$ & $TO$ && $TO$ \\
   \Xhline{0.5\arrayrulewidth}
   25 & 1566 & 2006 & 2100 & 1077 & \textbf{1318} & 1638 & 1093 & 7406 & 2043 & 1021 & 977 & 4465 && 1503 \\
   \Xhline{0.5\arrayrulewidth}
   26 & 62399 & 1886 & 62606 & 1802 & \textbf{61531} & 1939 & 62470 & 1908 & 62140 & 1807 & 60337 & 1867 && 1816\\
   \Xhline{0.5\arrayrulewidth}
   27 & 5318 & $TO$ & 27560 & 2603 & \textbf{9103} & 8188 & $166927^*$ & $TO$ & 28023 & 1621 & 8096 & 8430 && $TO$\\
   \Xhline{0.5\arrayrulewidth}
   28& 51638 & 1757 & 51803 & 1696 & \textbf{49528} & 1779 & 51764 & 1821 & 50893 & 1654 & 50765 & 1767 && 1684 \\
   \Xhline{0.5\arrayrulewidth}
   29 & 49109 & 5880 & 81882 & 5545 & \textbf{42791} & 5671 & 62527 & 6848 & 83461 & 5502 & $60410^*$ & 5802 && 49407\\
   \Xhline{0.5\arrayrulewidth}
   30 & 8972 & $6687^*$ & 8333 & $6518^*$ & \textbf{8228} & 8580 & 8805 & 6366 & 8227 & $6468^*$ & 8091 & 8516 && 8863\\
   \Xhline{0.5\arrayrulewidth}
   31 & 778 & 745 & 817 & 741 & \textbf{719} & 711 & $764^*$ & 1027 & 6150 & 761 & 757 & 1082 && 721\\
   \Xhline{0.5\arrayrulewidth}
   32 & 457 & $444^*$ & 411 & $410^*$ & \textbf{422} & 447 & 468 & 48481 & 414 & $472^*$ & 453 & $48076^*$ && 412 \\
   \Xhline{0.5\arrayrulewidth}
   33 & 287 & 41545 & 3274 & 7987 & \textbf{316} & 22566 & 298 & 42172 & 3177 & 7735 & 324 & 22407 && 308 \\
   \Xhline{0.5\arrayrulewidth}
   34 & 254 & 264 & 239 & 224 & \textbf{243} & 286 & 235 & 8267 & 248 & 236 & 283 & 8600 && 255\\
   \Xhline{0.5\arrayrulewidth}
   35 & 5717 & 5722 & 6744 & 4897 & \textbf{12081} & 4966 & 7413 & 9278 & 6769 & 4958 & 23110 & 7328 && 5378
 \\
   \hline
\end{tabular}
\end{adjustbox}
\end{sidewaystable}

\setcounter{topnumber}{3}

\begin{table}[t]
   \caption{F-score of test data from ORE 2014} 
   \label{tab:fscore}
   \centering % center the table
   \begin{tabular}{r|c|c|c|c|c|c|c|c|c|c|c|c} % alignment of each column data
   Config\# & 1 & 2 & 3 & 4 & 5 & 6 & 7 & 8 & 9 & 10 & 11 & 12 \\ 
   \hline
   \hline
   F-score & 94\% & 84\% & 94\% & 92\% & 96\% & 92\% & 88\% & 92\% & 92\% & 93\% & 94\% & 91\%
   \end{tabular}
\end{table}

A total of 35 ontologies have been chosen as test data (see Table \ref{tab:test}). The F-Score for each configuration on all 35 test data is given in Table \ref{tab:fscore}.

Table \ref{tab:test} demonstrates the satisfiability testing runtimes of all 35 tests on all of the heuristics. The last column shows the satisfiability testing runtimes of the tests on JFact Standard where the default heuristics are chosen. The symbol $^*$ means that a label is falsely predicted. For example, $4201^*$ indicates that the predicted label is ``Bad'' where the actual runtime is ``Good''.

If there are multiple ``Good'' choices, the ``Good'' with the highest priority will be chosen. For example, in sample 5, the configurations 2,4,6 and 10 are ``Good''. Since among these configurations, configuration 10 has the highest priority from Table \ref{tab:priority}, it will be chosen as the right heuristic. 

In a case where all of the labels of configurations are ``Bad'', the label with the lowest priority will be chosen with the hope for a false ``Bad'' prediction.

As shown from Table \ref{tab:test}, there are 6 cases (samples 5, 11, 20, 22, 24 and 27) in which our learning model outperformed the standard JFact by 1 or 2 orders of magnitude.

\begin{table}[t]
   \caption{Speedup factor of test data} 
   \label{tab:speedup}
   \centering % center the table
   \begin{tabular}{r|c|c} % alignment of each column data
   Speedup ratio & Timeout (500,000 ms) & Timeout (1,800,000 ms) \\ 
   \hline
   \hline
   Maximum & 167 & 602 \\
   Average & 11.6 & 39.56
   \end{tabular}
\end{table}

\begin{table}[t]
   \caption{Average and sum runtime (in milliseconds) of all test data with timeout=1,800,000 ms} 
   \label{tab:runtime}
   \centering % center the table

   \begin{tabular}{r|c|c|c} % alignment of each column data
    & JFact Learner & JFact Standard\\ 
      \hline
      \hline
   Sum & 2,741,960 & 13,604,733 \\ 
     \hline
   Avg & 78,341 & 388,707 

   \end{tabular}
\end{table}

Table \ref{tab:speedup} shows the average as well as maximum speedup of the satisfiability testing on the learned method compared to standard JFact. If the timeout is set to 1,800,000 milliseconds, the average speedup increases from 11.6 to 39.56 and the maximum from 167.28 to 602.208. Moreover, Table \ref{tab:runtime} shows the average and sum of the runtimes for our learning-based JFact is 4 to 5 times smaller compared to standard JFact; that is a significant improvement.

The runtime overhead for measuring the features varies from 5 to 15 seconds or could exceed this amount for very large-size ontologies. Since some of these features are related to the \texttt{DAG} structure obtained at the beginning of JFact, a small part of this time is due to the interoperation or switching time between Python (machine learning source code) and Java (reasoner source code) programs. However, this time can be reduced in future developments by optimizing the implementation. The other part of the overhead is due to the calculation of metric features related to the ontologies' structure that will take only a couple of seconds for smaller-sized ontologies (with a size less than 30 MB), but for large-sized ontologies that require more than 30 MB, this could take somewhere between 15 to 50 seconds. The number of very large-size ontologies makes up only 6\% of our data. 

Besides, almost all of these structural features are the same metrics that are measured\footnote{The metrics can be found in the metric section in the home display of Protege} at the beginning of Protege\footnote{\url{http://protege.stanford.edu}} framework after loading an ontology. The loading time could vary in Protege depending on the size of the ontology. For developing the built-in version of the learned based reasoner for Protege, these features are already measured in Protege while loading ontologies, i.e., there is no need to calculate them again.

\section{Conclusions and Future Work}
OWL reasoners are provided with many heuristic-based optimization techniques that can benefit them significantly if chosen properly. The optimization technique that is covered in this paper is the expansion ordering heuristics for satisfiability tests. These heuristics determine the order of concepts or sub-concepts for satisfiability tests. Selecting the right ordering helps to speed up reasoners for satisfiability testing and many other reasoning tasks that contain satisfiability testing.

Our learning-based method can further be improved in a way that it becomes closer to selecting not only the right but the best possible heuristic as well. This can be done by defining more thresholds that lead to determining more than ``Good'' and ``Bad'' labels. Although, as mentioned in Section \ref{sec:experiments}, we attempted to provide a solution to this by defining ``Ok'' labels, but maintaining a high accuracy model that holds the prediction of more than two labels is difficult. Therefore, building a high accuracy model that can reach this goal is highly appealing.

Further, a learning model may estimate the feature calculation time for very large-sized ontologies and predict whether it is worth for a reasoner to ignore and not apply the learning technique on those ontologies.

More reasoning tasks such as classification can be sped up that considers the combination of expansion ordering heuristics for satisfiability and subsumption testing together.

Finally, we believe there exist other optimization techniques developed for JFact or other DL reasoners that can be enhanced by the use of machine learning.

\clearpage
\bibliographystyle{splncs04}
\bibliography{paper}

\end{document}